\documentclass[10pt,twocolumn,letterpaper]{article}

\usepackage[pagenumbers]{cvpr}      
\usepackage{bm}
\usepackage{tikzducks}
\usepackage{graphicx}
\usepackage[most]{tcolorbox}
\usepackage{colortbl}
\usepackage{pifont}
\usepackage{enumitem}
\definecolor{Gray}{gray}{0.9}
\usepackage{amssymb}
\usepackage{pifont}
\newcommand{\cmark}{\ding{51}}%
\newcommand{\xmark}{\ding{55}}%
\usepackage[linesnumbered,ruled,vlined]{algorithm2e}
\newcommand\blfootnote[1]{%
  \begingroup
  \renewcommand\thefootnote{}\footnote{#1}%
  \addtocounter{footnote}{-1}%
  \endgroup
}

\usepackage{xcolor}

\definecolor{cvprblue}{rgb}{0.21,0.49,0.74}
\usepackage[pagebackref,breaklinks,colorlinks,citecolor=cvprblue]{hyperref}

\def\paperID{1447} 
\def\confName{CVPR}
\def\confYear{2024}

\title{\emph{V}$^*$: Guided Visual Search as a Core Mechanism in Multimodal LLMs}

\author{Penghao Wu \textsuperscript{$\dagger$}\\
UC San Diego\\
{\tt\small pew011@ucsd.edu}
\and
Saining Xie\\
New York University\\
{\tt\small saining.xie@nyu.edu}
}

\begin{document}

\twocolumn[{
\maketitle
\vspace{-30pt}
\begin{center}
    \includegraphics[width=0.95\linewidth]{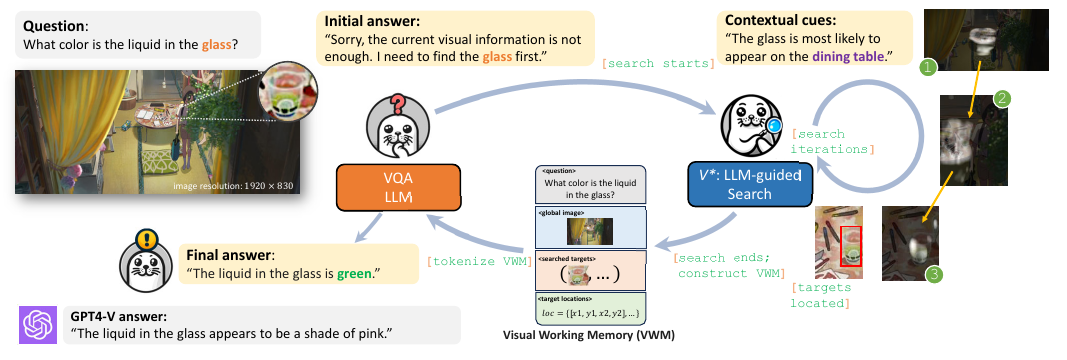}
    \captionof{figure}{The visual search mechanism enables humans to identify a target within a multitude of stimuli, streamlining the organization of information critical for problem-solving and reasoning. In this work, we explore this core mechanism in the context of MLLMs, addressing its absence, which currently impedes precise visual grounding, especially for high-resolution images. In this example, the VQA LLM could not immediately answer the question, thus activating \emph{V$^*$}, an LLM-guided visual search process that uses common sense and contextual cues to search for the required details. Throughout this informed search, it builds a visual working memory (VWM), tokenizing the overall context and the areas of interest related to the targets, which are then re-fed to the VQA LLM, enabling it to accurately answer the question.}
    \label{fig:teaser}
\end{center}
}]

\blfootnote{\textsuperscript{$\dagger$} Work done during an internship at NYU.}

\begin{abstract}
\vspace{-1em}
When we look around and perform complex tasks, how we see and selectively process what we see is crucial. However, the lack of this visual search mechanism in current multimodal LLMs (MLLMs) hinders their ability to focus on important visual details, especially when handling high-resolution and visually crowded images. To address this, we introduce V$^*$, an LLM-guided visual search mechanism that employs the world knowledge in LLMs for efficient visual querying. When combined with an MLLM, this mechanism enhances collaborative reasoning, contextual understanding, and precise targeting of specific visual elements. This integration results in a new MLLM meta-architecture, named \textbf{S}how, s\textbf{EA}rch, and Tel\textbf{L} (SEAL). We further create V$^*$Bench, a benchmark specifically designed to evaluate MLLMs in their ability to process high-resolution images and focus on visual details. Our study highlights the necessity of incorporating visual search capabilities into multimodal systems. The code is available \href{https://github.com/penghao-wu/vstar}{here}.

\end{abstract}    
\section{Introduction}
\label{sec:intro}

One of the hallmarks of human intelligence is being able to process and integrate multi-sensory information to perform complex tasks. A salient aspect of our cognitive reasoning process involving visual information is the ability to conduct \textit{visual search} – the process of efficiently recognizing and localizing key objects within intricate real-world scenes. This mechanism plays a fundamental role in the interaction with the environment and happens everywhere, from finding keys on a cluttered table to searching for a friend in the crowd. Besides, it is also an indispensable step for complex tasks that require multiple reasoning steps. The intricacy of visual search has been studied for a long time in cognitive science and vision science \cite{torralba2006contextual,peelen2011neural,wolfe2011visual,wolfe2017five,wolfe2020visual,wang2023statistical}.

While visual search seems intuitive for humans, it is actually a complex process underpinned by a series of complex behaviors. To accomplish this task efficiently, top-down feature guidance and contextual scene guidance are two fundamental factors, guiding humans' visual search process \cite{wolfe2017five}. The top-down feature guidance directs humans' attention to items with specific features or attributes (\eg color, shape, and orientation) based on the specification of the target object or knowledge about its general category. The contextual scene guidance is based on the fact that objects are usually well-organized in structured scenes in real-world scenarios. Therefore, one can use the semantics of the scene, object co-occurrence, and other physical constraints based on common sense knowledge to pay attention to specific regions, accelerating the search process.

\begin{figure*}
\centering
\includegraphics[width=0.91\linewidth]{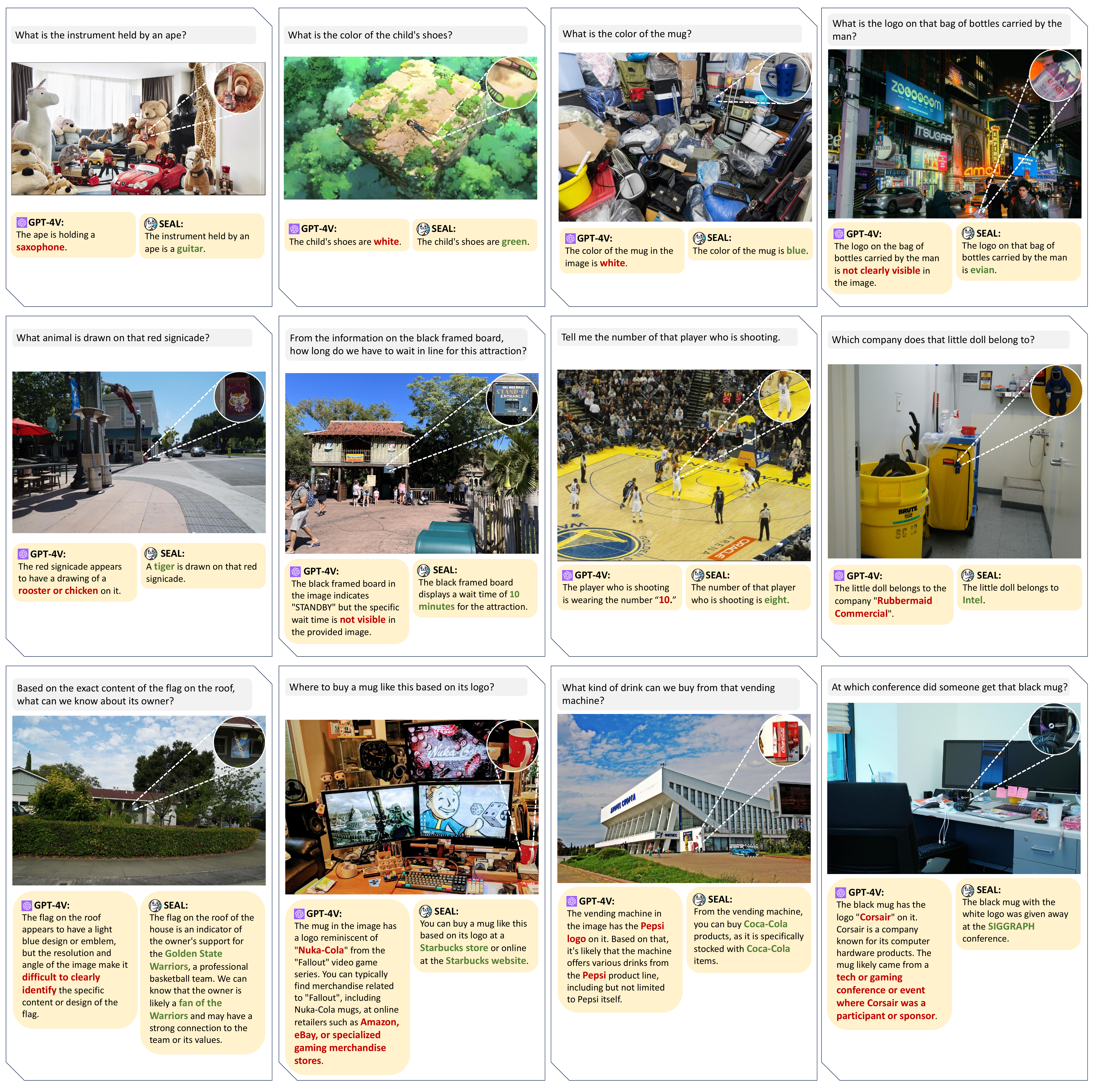}
\vspace{-0.5em}
\caption{Examples on which GPT-4V fails \text{\footnotesize{(\texttt{Accessed}: \texttt{Oct}\:\texttt{31},\:\texttt{2023})}} while SEAL with the \emph{V}$^*$ visual search mechanism succeeds. Even though GPT-4V has a much more powerful LLM (GPT-4) than ours (Vicuna-7B), it still occasionally struggles in scenarios that demand extensive visual processing. These situations require precise visual grounding in high-resolution images, a task where the visual search mechanism becomes essential. \emph{Best viewed on screen with zoom}. Image sources are provided in Appendix.}
\vspace{-1.5em}

\label{fig:comparison}
\end{figure*}

As an important step towards achieving artificial general intelligence, multimodal LLMs (MLLMs) \cite{blip2,flamingo,minigpt4,llava,InstructBLIP} try to emulate humans' ability to integrate multimodal information and perform general tasks. Significant advances have been made in this domain, leveraging the strong reasoning capabilities of large language models. However, a key limitation of current MLLMs is their dependence on pre-trained (and often frozen) vision encoders, such as the CLIP~\cite{CLIP} image encoder. This dependency forms a major bottleneck for visual information processing. 
The vision encoder is often trained on images with low resolution, such as 224$\times$224 or 336$\times$336 pixels. During deployment, images are also often resized to a lower resolution. As a result, the encoder may overlook important details in high-resolution images. Additionally, current MLLMs struggle to identify which essential visual details are missing or unclear in the images they process, nor can they proactively seek out or request this missing information.

Inspired by human capabilities, we propose \textbf{SEAL}
(\textbf{S}how, S\textbf{EA}rch, and Tel\textbf{L}), a general meta-architecture to integrate an LLM-guided visual search mechanism into MLLMs to address the aforementioned visual limitations. The SEAL framework consists of a VQA LLM and a visual search model. Unlike typical MLLM models that might refuse to answer or make uninformed guesses (\ie hallucinations) due to insufficient information from the vision encoder, the VQA LLM in SEAL can explicitly pinpoint the visual details that are missing, thus creating target objects for focus. Then, using the rich world knowledge and common sense of language models, the visual search component locates these identified elements, adding them to a Visual Working Memory (VWM). This additional visual data in the VWM enables the VQA Language Model to provide more accurate and informed responses. SEAL's adaptability allows it to work with various MLLM base models; in our case, we use LLaVA~\cite{llava} as both the VQA LLM and the MLLM in the visual search model. With this new visual search capability, the MLLM is better equipped to handle situations that require accurate visual grounding in high-resolution images, as highlighted in our comparison (Fig~\ref{fig:comparison}).

As humans' visual search process is guided by top-down feature guidance and contextual scene guidance, we design an informed visual search algorithm dubbed \emph{V}$^*$ with a visual search model following similar principles. For humans, such guidance largely comes from their knowledge and experiences about the physical world. Thus, our visual search model is built atop another MLLM which contains vast common sense knowledge about the world and can effectively reason about the possible locations of the target in the scene based on this knowledge. 

The existing MLLM benchmarks \cite{mme, mmbench, seed-bench} primarily focus on providing comprehensive evaluations across various task categories, and do not adequately challenge or expose the specific limitations of current paradigms mentioned above. To bridge this gap and evaluate our proposed framework, we introduce \textbf{\emph{V}$^*$Bench}, a new dedicated VQA benchmark that focuses on detailed visual grounding on high-resolution images. \emph{V}$^*$Bench is a vision-focused benchmark, requiring multimodal models to accurately ground specific visual information that could be easily overlooked by a standard, static vision encoder lacking visual search capabilities. In a world increasingly dominated by rich and complex visual content like images and videos, it's crucial for MLLMs to be able to actively focus on critical visual information for complex reasoning tasks. This benchmark aims to highlight the significance of this fundamental mechanism and guide the evolution of MLLMs towards mirroring the multimodal processing and reasoning aptitudes inherent in human cognition.

In summary, our contributions are threefold: 1) We propose SEAL, an MLLM meta-architecture designed to actively reason about and search for needed visual information, a vital capability for vision-intensive multimodal tasks, especially when dealing with high-resolution images. 2) We develop a visual search algorithm \emph{V}$^*$ that utilizes the common sense understanding inherent in LLMs to perform efficient informed searches across images of any resolution. 3) We introduce \emph{V}$^*$Bench to thoroughly evaluate the ability of MLLMs in accurately processing and grounding detailed visual information in high-resolution images.

\section{Related Work}

\subsection{Computational Models for Visual Search}

Inspired by guiding factors in humans' visual search process,  several computational models have been proposed to mimic the human visual search process. \citet{sclar2020modeling} proposes a Bayesian searcher combined with a saliency map as prior. 
\citet{torralba2006contextual} combines the local saliency map with the global scene priors to form a scene-modulated saliency map. IVSN \cite{zhang2018finding} uses convolutional networks to compute the similarity map between the search image and the target template and perform the search greedily. \citet{yang2020predicting} uses inverse reinforcement learning (IRL) to learn the reward function and policy of human visual search. 

Nevertheless, such models mainly focus on mimicking the human gazing trajectory, without requiring accurately localizing the target object. And they usually adopt a fixed-size gazing window while our visual search model tackles any resolution images in a hierarchical process. Besides, their usage of categorical information about the target objects and the contextual scene information is limited to simple statistics and does not generalize to general domains. Our visual search model utilizes the rich common sense knowledge from LLM to expedite the search process. 
We note that our active search strategy is linked to System II cognitive processes~\cite{fastslow} -- for complex tasks, dynamic computation allocation for visual search becomes necessary. Our approach can also be thought as a visual counterpart to the chain-of-thought (CoT) technique used in LLMs~\cite{cot}.

\vspace{-1mm}

\subsection{Multimodal LLMs}

Propelled by the success of large language models, vision language model research begins to explore how to equip LLMs with additional vision input to solve various multimodal tasks. Currently, MLLMs can be categorized into two types: end-to-end models and LLM tool-using systems.

\noindent \textbf{End-to-end MLLMs.}  End-to-end MLLMs\cite{flamingo, blip2, llava, minigpt4, InstructBLIP, otter} connect the pre-trained LLM with a vision encoder through projection or alignment modules, and the whole system is jointly trained in an end-to-end manner. These models aim to project the visual features to the input embedding space of language or intermediate feature space, enabling the LLM to process visual information and perform vision-language tasks. While vision encoders like CLIP~\cite{CLIP}, which are pre-trained through image-text alignment, can translate visual features into a form of `language tokens' understandable by LLMs, this process introduces an information bottleneck. The conversion and projection of visual features often lead to inherent information loss, especially since vision encoders are typically constrained to low-resolution images. Consequently, these models may struggle to provide accurate results or might produce hallucinated answers if crucial visual information is poorly captured or inadequately focused upon.

\noindent \textbf{LLM-tool-using systems.} LLM-tool-using systems or LLM-based agents treat the LLM as a black box and give them access to some vision expert systems to perform certain vision-language tasks through reasoning \cite{mm-react,  ChatCaptioner, visualChatGPT, chameleon, idealgpt, AVIS}. Such systems utilize different kinds of vision experts to provide needed information about the visual input in the form of \textbf{text}. They usually adopt captioning and detection models to create general textual information about an image which is then provided to the LLM. Based on the description of the image and a certain question or task instruction, the LLM further decides what visual information is needed and which visual experts to call through reasoning. 
The LLM decides to terminate the process and provide the final answer when it thinks the information is enough. However, one main problem of such systems is that as the whole system is running based on text only, certain visual information might be inevitably ignored or distorted when translated into text. Moreover, as the vision experts are not perfect themselves, cascaded errors exist and the complex and lengthy process makes the whole system prone to fail.

\section{Method}
\label{sec:method}

\begin{figure*}[ht]
\centering
\includegraphics[width=0.98\linewidth]{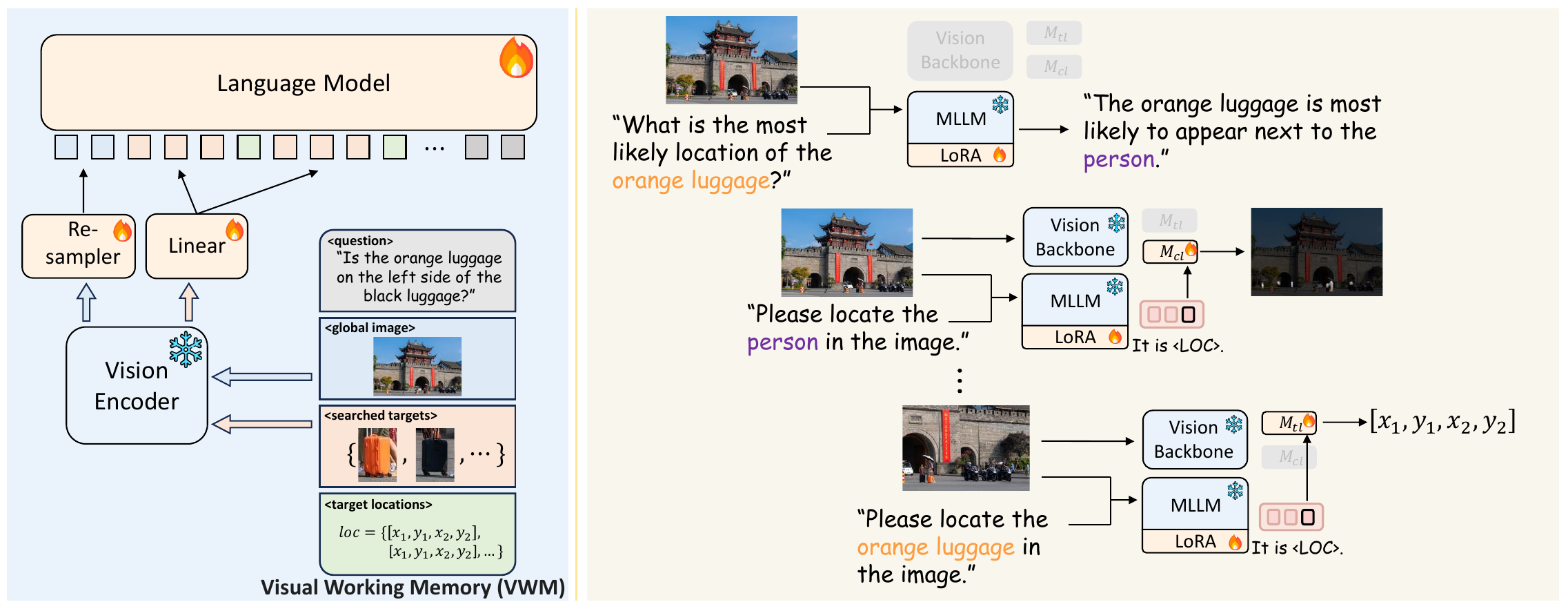}
\vspace{-0.5em}
\caption{An instantiation of the proposed SEAL framework. The left section represents the VQA LLM, which utilizes all the data within the Visual Working Memory to respond to questions. On the right, we illustrate the operational pipeline of the \emph{V}$^*$ visual search algorithm.}
\vspace{-1.5em}
\label{fig:framework}
\end{figure*}

Our proposed Show, Search and Tell (SEAL) framework is a general meta-architecture for MLLMs. It is comprised of a VQA LLM and a visual search model which collaborate and interact through the visual working memory (VWM). An illustration of the SEAL framework is shown in Fig~\ref{fig:framework}. In this work, we provide an instantiation of SEAL to validate its effectiveness and choose the LLaVA-7B model as the MLLM in the SEAL framework.
We now elaborate on the model structures of each of these two parts. The training data curation process for the visual search model and the training details are provided in the Appendix~\ref{appx:model_training}.
\subsection{VQA LLM with Visual Working Memory}
\subsubsection{Model Structure}
Modern MLLMs usually have three components: a vision encoder, a projection module, and an LLM. The type of projection module varies across different models. including options like Resampler \cite{flamingo,otter,wang2023makes}, QFormer \cite{blip2, InstructBLIP}, and linear layer \cite{llava, chen2023shikra}. 
The placement of the projected vision tokens within the LLM also differs among models, such as in the input layer \cite{blip2,InstructBLIP,minigpt4,llava}, or middle cross-attention layers \cite{flamingo,otter}. Despite these variations, most of these models adopt pre-trained CLIP as their vision encoder. When tackling high-resolution and visually-crowded images, the visual features extracted by CLIP may not capture the necessary information required to answer the question.

\begin{algorithm}
\scriptsize
\DontPrintSemicolon
\SetAlgoNoEnd
\caption{SEAL Working Pipeline}
\label{algo:SEAL}
\SetKwFunction{FMain}{SEALVQA}
\SetKwFunction{VS}{Visual Search}
\SetKwFunction{VQA}{VQALLM}
\SetKwProg{Fn}{Function}{:}{}
\Fn{\FMain{$I$, $T$, $\delta$}}{
    list of needed target objects $L$ $ \leftarrow$  \VQA{$I$, $T$}\;
    Initialize VWM \;
    VWM.add($I$),
    VWM.add($T$) \;
    \For{target in $L$}{
        Priority queue $q$\;
        $q.add(($I$, \infty))$\;
        search result $ \leftarrow$ \VS{$q$, $s$, $\delta$}\;
        \If{search result is None}{
            VWM.add("\{target\} not existent in the image")
        }
        \Else{
            Cropped the object patch from $I$\;
            VWM.add("\{target\} $<$object patch$>$ at location $[x1, y1, x2, y2]$")\;
        }
    }
    response $ \leftarrow$  \VQA{VWM}\;
    \KwRet response\;
}    
\end{algorithm}

The visual search mechanism is not always engaged. The model first evaluates if the encoder's initial (global) visual features suffice for answering the question. If not, it explicitly lists all the needed but missing information in the format of a list of target objects. Then, it initializes a visual working memory (VWM). The VWM has four blocks, the \texttt{$<$question$>$} block contains the initial textual question; \texttt{$<$global image$>$} contains the initial image; \texttt{$<$searched targets$>$} stores the target object crops after search; and \texttt{$<$target location$>$} stores the coordinates of the searched targets. Next, the visual search model searches over the image and localizes each required target. A region containing the identified target is then cropped from the whole image. The cropped targets, along with their coordinates, are added to the VWM. After that, the VQA LLM processes the data contained in the VWM to generate the response accordingly. The working pipeline of the SEAL framework is illustrated in Algorithm~\ref{algo:SEAL}.

 In this work, we choose the CLIP ViT-L/14 model~\cite{CLIP} as the visual feature extractor, with input resized and padded to $224^2$. We use it to process both the initial image and the crops of searched targets. To adapt the visual features for input into the LLM, we consider two types of projection modules, the linear layer and the resampler. The linear layer projection module keeps the number of visual tokens from the vision encoder, and the cross-attention based resampler projection reduces the number of tokens (\ie 256 to 32). To manage the token length corresponding to different contents in VWM, we have designed a simple scheme to flexibly switch between these two projection modules. In scenarios where the input comprises only the initial image feature without any searched targets, we apply the linear layer projection to maintain all visual tokens. When one or two searched targets are present in the VWM, it's presumed that the model needs to focus on these targets. In such cases, we use the linear layer projection for the visual features of these targets and employ the resampler to subsample the global image features. For situations where the VWM holds more than two searched targets, the resampler is used for all visual features to reduce computational cost.
 
\subsection{Data Curation for VQA LLM}
Since our VQA LLM will now work with the VWM that has searched targets, we need to perform additional instruction tuning to train the VQA LLM. We describe the training data as follows. More details can be found in the Appendix~\ref{appx:data_vqa_llm}.

\noindent\textbf{Negative data for target objects reasoning (100k)} The VQA LLM must first identify the target objects that are 1) required to answer the question, and 2) missing or not clear enough in the initial global image features. To facilitate this, we construct (image, question, answer) data where the question pertains to one or two objects not present in the image. Additionally, we construct questions about details of certain objects, deliberately made too small to be captured by the CLIP encoder. This is achieved by choosing objects with bounding box sizes smaller than $20\times20$. The appropriate response to such questions is a straightforward acknowledgment that the question cannot be answered, along with a clear enumeration of all the additional target objects required.  We construct 100k data on COCO2017 \cite{COCO} with questions generated by GPT-3.5.

\noindent\textbf{VQA data (167k)} This data consists of three parts: GQA data (70k) from \cite{GQA}, VQA data focused on object attributes (51k), and VQA data focused on spatial relationship (46k).
In the GQA subset, we utilize the original dataset's GT annotations about specific objects mentioned in the questions. We select a portion of this data, treating the mentioned objects as search targets in the VWM during training. Additionally, we rephrase the short answers in GQA into full sentences using GPT-3.5. For the object attribute data, we utilize the VAW~\cite{vaw} data, transforming them into question-answer pairs in a standard format that inquire about certain object attributes, and consider these objects as search targets. 
Regarding the spatial relationship data, we use the COCO2017 dataset to generate questions about the relative spatial positioning of two objects within an image, treating these two objects as the search targets.

\noindent\textbf{LLaVA Instruction Tuning (120k)} To maintain the general multimodal question answering and instruction following capabilities, we also include the LLaVA-80K instruction tuning data, of which the image sources are also COCO. Additionally, we identify object entities in the questions that match with COCO categories and have box annotations. These matched objects are then designated as the search targets, creating an additional set of 40k data.

\subsection{\textbf{\emph{V}}$^*$: LLM-guided Visual Search}
\subsubsection{Problem Formulation}
At a high level, the objective of visual search shares similarities with the task of referring expression comprehension (REC)~\cite{mao2016generation} in computer vision. REC aims to localize a specific object in the image as described by a textual referring expression. However, unlike REC, which is restricted to images of a specific size, visual search must adapt to images of any resolution. Sometimes, a thorough search across the entire image is needed to find the target object. Consequently, \emph{visual search efficiency matters}: an effective visual search algorithm should not only locate the target accurately but also do so as quickly as possible.

\subsubsection{Model Structure}
Similar to how people often zoom in on their phones for a clearer view, when dealing with a high-resolution image, it's possible that the target object cannot be precisely identified and located if only the entire image is viewed as a small thumbnail. To address this, one straightforward approach is to patchify an image into uniformly sized small patches and perform the localization on each patch exhaustively. This brute-force strategy is often used in aerial image detection and whole slide image analysis~\cite{ozge2019power, chen2022scaling}. However, it tends to be too inefficient for effectively managing images with very high resolutions -- we need a smarter solution.

Drawing inspiration from how humans utilize contextual scene and top-down feature guidance in their visual search process, we've incorporated similar concepts into the design of the visual search model in \emph{V}$^*$. This process utilizes an MLLM that encapsulates a vast amount of common sense knowledge, serving as heuristic guidance. In order to localize and crop the searched targets for VWM, it's also necessary to enhance the MLLM with additional localization capabilities, comparable to those mentioned in \cite{LISA, contextDET}.

\begin{figure}
\centering
\includegraphics[width=\linewidth]{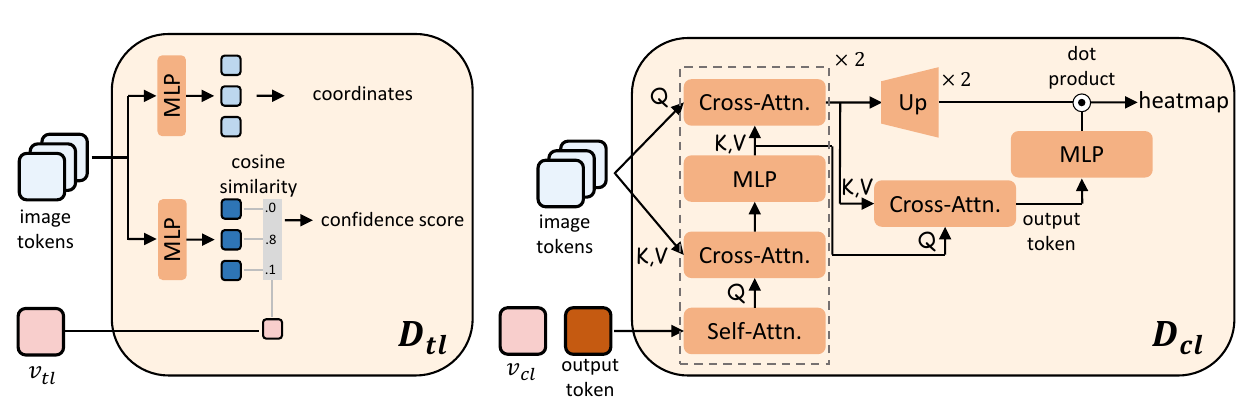}
\caption{Detailed structure of the \emph{target} localization decoder $D_{tl}$, and the \emph{search
cue} localization decoder $D_{cl}$.}
\vspace{-1.5em}
\label{fig:decoder}
\end{figure}

Our visual search model consists of an MLLM and a localization module with an image backbone and two decoders, \ie a target localization decoder $D_{tl}$ and a search cue localization decoder $D_{cl}$. The MLLM has an additional localization ability with a localization token \texttt{<LOC>} added to its vocabulary. Given an image and a textual expression of an object or region, the textual expression is first transformed into a fixed-format instruction (\ie ``Please locate the \texttt{[object]} in the image.") and then fed into the MLLM together with the image. The MLLM outputs the localization token \texttt{<LOC>} containing contextual and location-related information of the queried textual expression. We process the \texttt{<LOC>} token embedding $\bm{v}_{loc}$ with two separate MLPs to get two additional embeddings $\bm{v}_{tl}$ and $\bm{v}_{cl}$. The image tokens from the visual encoder are then combined with $\bm{v}_{tl}$ and $\bm{v}_{cl}$, processed by decoders $D_{tl}$ and $D_{cl}$ respectively, and output \emph{target coordinates} (with confidence scores) and \emph{search cue heatmap} respectively. The $D_{cl}$ resembles the mask decoder in SAM~\cite{SAM}
, and the $D_{tl}$ is implemented with two linear heads, one for coordinate prediction and the other for confidence score prediction. The detailed structure of these two modules is shown in Fig~\ref{fig:decoder}. 

\subsubsection{Search Algorithm}

With this visual search model, our \emph{V}$^*$ algorithm works as follows. Given an image and a textual expression of the target object, the \emph{V}$^*$ MLLM first attempts to locate the target directly. In this step, we obtain the \emph{target coordinates} and the \emph{search cue heatmap} from $\bm{v}_{loc}$ corresponding to the target object. When no object is located (\ie the confidence score falls below a threshold), we examine the heatmap for possible target-specific cues. 

The search cue heatmap highlights regions that could potentially contain the queried target object. When the \emph{target-specific cue} is prominent (\ie when the highest value in the heatmap exceeds the threshold $\delta$), we use it to guide the search directly. Otherwise, we ask the MLLM what is the most likely location of the target object in the image. This requires the MLLM to utilize its common sense knowledge and integrate it with the image's context to provide the \emph{contextual cue} about the target's whereabouts. Upon receiving a description of the region where the target object is likely located, we then prompt the MLLM to locate the described area with the $D_{cl}$ decoder and produce a search cue heatmap corresponding to the contextual cue. 

\begin{figure}
\centering
\includegraphics[width=1.0\linewidth]{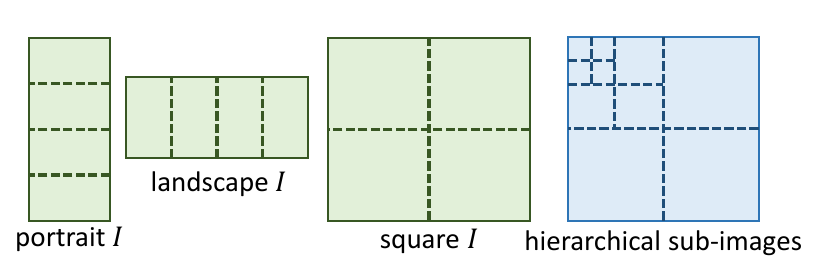}
\caption{Images are recursively divided into four patches based on their aspect ratio. Landscape images are divided vertically. Portrait images are divided horizontally.}
\vspace{-0.0em}
\label{fig:patch}
\end{figure}

Then, we use a simple strategy and recursively divide the image into 4 non-overlapping equal-sized patches.\footnote{A corner case is that this simple strategy might fail when targets are located at the boundaries of patches. One can use overlapping patches or variable-sized patches based on the heatmap distribution, if necessary.} In order to maintain a square-like aspect ratio for each patch during the search, we adjust our division strategy based on the image's orientation. For landscape images (\ie, where width is greater than twice the height), we divide the image vertically. Conversely, for portrait images (\ie, where height exceeds twice the width), we divide it horizontally. In all other cases, we split the image both horizontally and vertically. This approach to patching is depicted in Fig~\ref{fig:patch}. Subsequently, we assign \emph{search priority scores} to these patches. The search priority score is calculated from the search cue heatmap (either target-specific or contextual). Based on the priority scores, the patches are then cropped and processed sequentially. This recursive procedure is repeated until the target object is located or the size of the current patch becomes smaller than a predetermined threshold. The overall process of the \emph{V}$^*$ algorithm is illustrated in Algorithm~\ref{alg:visual search}.

\noindent\textbf{Connection to \emph{A}$^*$ Algorithm.} The naming of our LLM-guided visual search \emph{V}$^*$ algorithm is inspired by its similarities to the informed search algorithm \emph{A}$^*$. \emph{A}$^*$ is designed for pathfinding, aiming to identify the shortest route between a starting point and a goal by using a heuristic to approximate the cost. In the context of our LLM-guided visual search, \emph{V}$^*$ can be seen as a unique variant of \emph{A}$^*$, where sub-images are treated as nodes. The cost function  $g(n)$ is set as a uniform positive constant for all nodes $n$, and the heuristic function $h(n)$ is defined as the negative of the priority score derived from the search cue heatmap. While the \emph{A}$^*$ algorithm's objective is to find a path with minimal cost from start to goal, our focus with \emph{V}$^*$ is solely on minimizing the total number of steps required to locate the goal.

\begin{algorithm}
\DontPrintSemicolon
\SetAlgoNoEnd
\scriptsize
\caption{\emph{V}$^*$: LLM-guided Visual Search}\label{alg:visual search}
\SetKwFunction{FMain}{Visual Search}
\SetKwFunction{VSM}{VisualSearchModel}

\SetKwProg{Fn}{Function}{:}{}
\Fn{\FMain{$q$, $s$, $\delta$}}{
    Current image $I_p$ = $q.pop()$\;
    target coordinates\&confidence, search cue heatmap $ \leftarrow$ \VSM{instruction="Please locate the $s$ in the image.", image=$I_p$}\;
    
    \If{target confidence is high}{ 
        \KwRet target coordinates\;
    }
    \If{heatmap.max() $< \delta$ }{
        contextual cue $ \leftarrow$ \VSM{instruction="What is the most likely location of the $s$ in the image?", image=$I_p$}
        \;
        \_ , search cue heatmap $ \leftarrow$ \VSM{instruction="Please locate the 'contextual cue' in the image.", image=$I_p$} \;
    }
    Divide $I_p$ into sub-images, calculate the priority for each sub-image based on the heatmap, and add (sub-image, priority) pairs to $q$. \;
    \While{$q$ is not empty}{
        search result $ \leftarrow$ \FMain{$q$, $s$, $\delta$} \;
        \If{search result is not None}{
            \KwRet search result\;
        }
    }
    \KwRet None
}
\end{algorithm}

\section{Benchmark}
\label{sec:benchmark}
To quantitatively evaluate MLLMs' ability in challenging scenarios where the image contains abundant and complex information and the visual information needed might not be easily found, we build a benchmark \textbf{\emph{V}$^*$Bench} based on 191 high-resolution images from SA-1B dataset~\cite{SAM} with an average image resolution of 2246$\times$1582.

Our benchmark contains two sub-tasks: attribute recognition and spatial relationship reasoning. The attribute recognition task has 115 samples and requires the model to recognize a certain type of attribute (\eg color, material) of an object. The spatial relationship reasoning task has 76 samples and asks the model to determine the relative spatial relationship between two objects. These tasks focus on evaluating the detailed visual analysis capability of the multimodal models. Both the test images and questions have been carefully selected and crafted by human annotators to ensure that it is difficult to directly ``guess'' the correct answer without accurate visual grounding of the relevant objects in the image. Examples of our benchmark can be found in the Appendix~\ref{appx:benchmark}.

For a quantitative comparison of open-source MLLM models on our benchmark, we construct multiple choice options for each question. We formulate four options for open-ended questions and two for binary questions. To ensure clarity, these multiple choices are carefully crafted and reviewed by human annotators for any potential ambiguity.
\section{Experiments}
\label{sec:experiments}
\subsection{Evaluation on \emph{V}$^*$ Bench}

\begin{table}[t]
\centering
\scalebox{0.85}{
\begin{tabular}{rccc}
\toprule
& Attribute (\%) & Spatial (\%)   & Overall (\%)   \\ \midrule
\rowcolor{Gray}
Human & 98.26 & 100.00 & 98.95  \\ 
\rowcolor{Gray}
Random Guess & 26.73 & 50.00 & 35.99  \\ \midrule
\multicolumn{4}{c}{\emph{Open-source end-to-end MLLMs}}  \\   
BLIP2 \cite{blip2} & 26.95 & 53.94 & 37.69   \\
MiniGPT-4 \cite{minigpt4} & 30.43 & 50.00 & 38.22  \\
LLaVA \cite{llava} & 23.47 & 53.94 &  35.59  \\
InstructBLIP \cite{InstructBLIP} & 25.21 & 47.36 & 34.02  \\
Otter \cite{otter} & 26.95 & 56.57 & 38.74  \\ 
LLaVA-1.5 \cite{llava1.5} & 43.47 & 56.57 & 48.68 \\
\midrule
\multicolumn{4}{c}{\emph{LLM tool-using pipelines}}  \\  
MM-React \cite{mm-react} &34.78 &51.31  &41.36 \\
VisualChatGPT \cite{visualChatGPT} & 30.43 & 48.68 & 37.69  \\
Visprog \cite{visprog} & 31.30 & 56.57 & 41.36  \\ \midrule
\multicolumn{4}{c}{\emph{Commercial chatbot systems}}  \\ 
Bard \cite{bard} & 31.30 & 46.05 & 37.17   \\
Gemini Pro \cite{Gemini} & 40.86  & 59.21 & 48.16 \\
GPT-4V \cite{gpt_4}& 51.30 & 60.52 & 
 54.97 \\ \midrule
SEAL (Ours) & \textbf{74.78} & \textbf{76.31} & \textbf{75.39}  \\ \bottomrule
\end{tabular}}
\caption{Evaluation of multimodal systems on \emph{V}$^*$Bench. We find our SEAL model outperforms leading-edge systems such as GPT-4V and Gemini by a large margin, even though we only use a Vicuna-7B LLM. This result demonstrates the importance of integrating a visual search mechanism into MLLMs.}
\vspace{-0.0em}
\label{Table:benchmark}
\end{table}

\begin{table*}[h]
\centering
\scalebox{0.9}{
\begin{tabular}{ccclccc}
\toprule
Experiment ID & LLM & VWM & Search & Attribute & Spatial & Overall \\ \midrule
1&  Vicuna-7B   &\xmark&N/A&38.26&55.26&45.02\\
2&  Vicuna-7B   &\cmark&Querying Detection (GD \cite{groundingdino})&62.60& 61.84& 62.30\\
3&  Vicuna-7B   &\cmark&Querying Detection (OWL-ViT \cite{owl-vit})&60.86&65.78& 62.82\\
4&  Vicuna-7B   &\cmark& \emph{V}$^*$-Search &\textbf{74.78}& \textbf{76.31}&\textbf{75.39}\\ \bottomrule
\end{tabular}}
\vspace{-0.5em}
\caption{Ablation studies on the necessity of the visual search mechanism. The LLaVA* denotes our VQA model without the visual search mechanism. Detection (GD) and (OWL-ViT) denote replacing the visual search model with GroundingDINO and OWL-ViT respectively.}
\vspace{-1em}
\label{Table:ablation}
\end{table*}

\begin{table}[h]
\centering
\scalebox{0.9}{
\begin{tabular}{cc}
\toprule
& Search Length $\downarrow$ \\ \midrule
Random-DFS     &    8.94           \\
Random-BFS     &    7.18           \\
Sequential-DFS &    11.39          \\
Sequential-BFS &    6.62           \\
\midrule
LLM-guided visual search &  \textbf{4.65} \\  
w/o target-specific cue &   5.22 \\
w/o contextual cue     & 5.36 \\
\bottomrule
\end{tabular}}
\caption{Evaluation of different search strategies on \emph{V}$^*$Bench.}
\label{Table:det vs search}
\end{table}

\begin{figure}[h]
\centering
\includegraphics[width=0.9\linewidth]{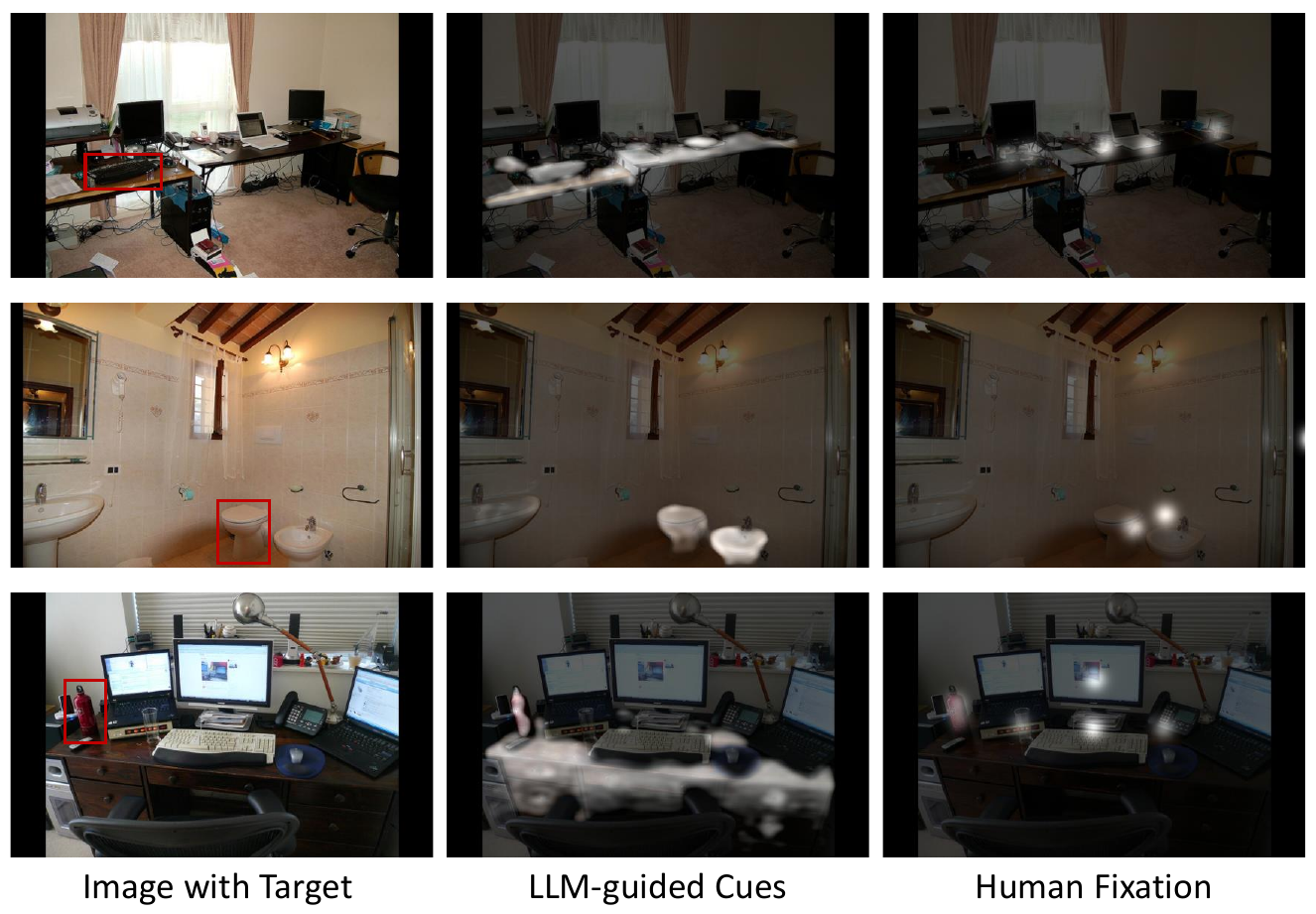}
\caption{Comparison with the human fixation on COCO-Search18~\cite{cocosearch18}. Humans tend to focus on center regions or salient objects while our model focuses on a larger contextual region.}
\vspace{-2mm}
\label{fig:cocosearch}

\end{figure}

\begin{table}[h]
\centering
\scalebox{0.9}{
\begin{tabular}{cc}
\toprule
               & Search Length $\downarrow$ \\ \midrule
Random-DFS     &  9.97             \\
Random-BFS     &  4.90             \\
Sequential-DFS &  9.82             \\
Sequential-BFS &  4.20             \\
Human Fixation ($\gamma$=0.9) &  2.52  \\
Human Fixation ($\gamma$=0.8) &  2.70 \\ \midrule
LLM-guided visual search &     2.80       \\ \bottomrule
\end{tabular}}
\caption{Comparison with the human fixation on COCO-Search18.}
\vspace{-1.5em}
\label{Table:cocosearch}
\end{table}

In this work, we implement the VQA LLM in our SEAL framework with Vicuna-7B\cite{vicuna} as the language model. We evaluate it with other open-source end-to-end MLLMs and LLM-tool-using systems on the proposed \emph{V}$^*$Bench. For end-to-end models, we include representative methods including \cite{blip2,minigpt4,llava,InstructBLIP,otter,llava1.5} and use the likelihood approach to evaluate their performance following \cite{GPT3, seed-bench}---we select the choice with the highest log-likelihood as the model's prediction. For the LLM-tool-using systems, we evaluate methods including MM-React \cite{mm-react}, VisualChatGPT \cite{visualChatGPT}, and Visprog \cite{visprog}. Additionally, we also evaluate industrial multimodal chatbots: Bard \cite{bard}, Gemini Pro \cite{Gemini}, and GPT4-V \cite{gpt_4}. For the LLM-tool-using systems and the multimodal chatbots, we prompt them to directly answer the option as the likelihood does not apply to them, and we ask them to choose the most likely option when they find that it is impossible to answer the question or none of the option is correct. We evaluate Bard and GPT4-V through the web chatbot (\texttt{Accessed}: \texttt{Oct}\:\texttt{31},\:\texttt{2023}) and evaluate the Gemini Pro through the API  (\texttt{Accessed}: \texttt{Dec}\:\texttt{16},\:\texttt{2023}).

As shown in Table~\ref{Table:benchmark}, we can see that the performance of most MLLMs is merely close to random guessing. The GPT-4V and Gemini systems can better handle some relatively easy scenarios in the attribute recognition task, but the overall performance is still not satisfactory. It's also noteworthy that the LLaVA-1.5 model, compared to the initial LLaVA model, shows a significant improvement in the attribute recognition task. This enhancement could be partially attributed to the adoption of a new vision encoder with a higher training resolution (CLIP-ViT-L-336px). However, there is still a considerable gap in performance when compared to our visual search strategy. With the visual search process, our model greatly improves performance. Nonetheless, considering that humans can achieve near-perfect results, there remains considerable potential for further improvement for MLLMs. Our visual search incurs an average time cost of 6.0 seconds per target on one A100 GPU. This is a reasonable trade-off, as, akin to human visual search and reasoning, allocating more computational resources is necessary for tackling challenging tasks.

\subsection{Ablation Study}
We conducted ablation experiments to verify the effectiveness of our key designs. First, we start with the LLaVA model that has the same structure (without the VWM) as our VQA LLM and train it on the same training data. Then, we replace the visual search mechanism with open-world detectors GroundingDINO \cite{groundingdino} and OWL-ViT \cite{owl-vit} and use the detection results to fill in the VWM. The experiment results are shown in Table~\ref{Table:ablation}. We can see that, although we include attribute recognition and spatial relationship reasoning data in the VQA LLM training data, the MLLM without the visual search mechanism (ID 1) still struggles. 
Direct querying detection models (ID 2\&3) as a substitute for the search process also results in significantly inferior performance compared to \emph{V}$^*$. Moreover, off-the-shelf detectors would encounter practical difficulties when applied to images of very high resolution.

\subsection{Visual Search Evaluation}

\newcolumntype{g}{>{\columncolor{Gray}}c}
\newcolumntype{e}{>{\columncolor{green!25}}c}
\newcolumntype{r}{>{\columncolor{red!25}}c}
\begin{table*}[h]
\centering
\scalebox{0.85}{
\begin{tabular}{ceeeggrr}
\toprule
 & \emph{V}$^*$Bench & MME & POPE & MMBench & SEED-Bench(Img) & MM-Vet & LLaVA$^{\rm W}$ \\ \midrule
 LLaVA* (7B) &45.0 & 1051.2 & 76.5 & 34.4 & 41.8 & 30.4 & 62.6 \\
 SEAL (7B) &  75.3  (+30.30) &  1128.9 (+77.70) &82.4 (+5.85) & 33.1 (-1.36) &  41.7 (-0.17) &  27.7 (-2.70) &  59.1 (-3.50)\\ \bottomrule
\end{tabular}}
\vspace{-0.5em}
\caption{When tested on a broader range of multimodal benchmarks, the addition of the visual search module mostly maintains the overall multimodal capability, and enhances performance in object hallucination benchmarks like POPE~\cite{POPE}.}
\vspace{-1em}
\label{Table:general benchmark}
\end{table*}

\begin{figure*}
\centering
\includegraphics[width=0.92\linewidth]{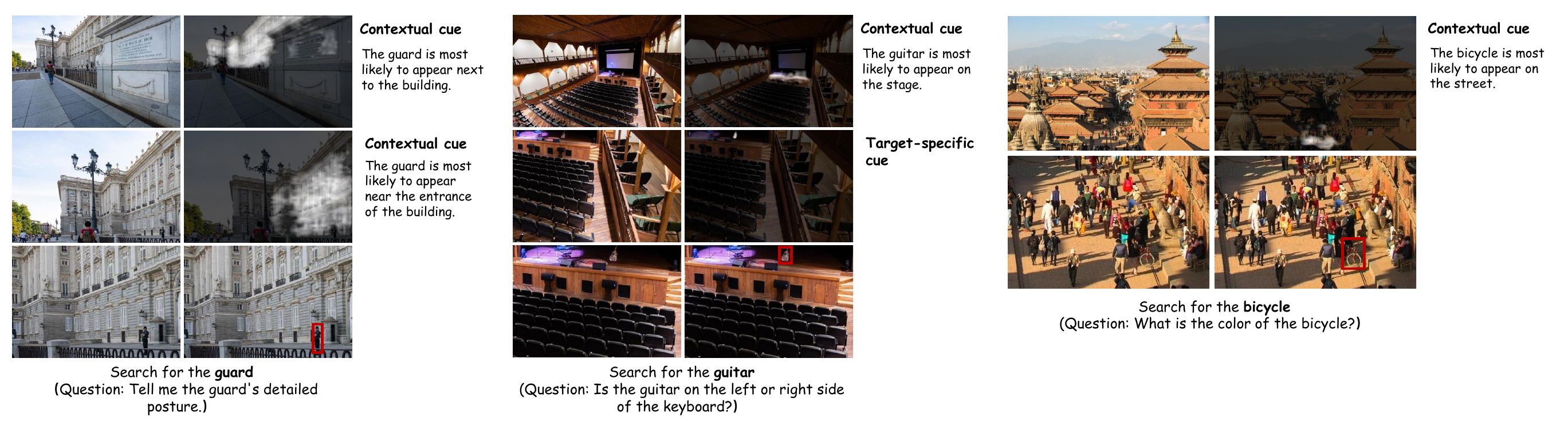}
\vspace{-1em}
\caption{Examples of the LLM-guided visual search process. Each row in each example represents a step in the visual search process and the heatmap of contextual cue or target-specific cue is shown on the right.}
\vspace{-1.5em}
\label{fig:search_process}
\end{figure*}

First, we record all 245 target object locations in the \emph{V}$^*$Bench. 
We then evaluate different search strategies in terms of search length. The search length here is defined as the number of search steps from the initial image to the patch where the target is located. We only include samples that can be successfully located after the search for evaluation. We compare our LLM-guided \emph{V}$^*$ algorithm with two baselines. The \emph{Random} baseline adopts the random search strategy that picks a random sub-image to explore, and the \emph{Sequential} baseline searches the sub-images sequentially, following a reverse raster scan order. These two strategies are evaluated in breadth-first search (BFS) and depth-first search (DFS) settings respectively. As shown in Table~\ref{Table:det vs search}, \emph{V}$^*$ could greatly reduce the average search length, and both the target and contextual search cues are helpful. We provide visualizations of the search process in Fig~\ref{fig:search_process}.

To further study the efficiency of \emph{V}$^*$ algorithm and draw parallels with cognitive science research in visual search, we conduct comparisons between our search outcomes and human behaviors using the COCO-Search18 dataset~\cite{cocosearch18}. COCO-Search18 records people's eye fixations when searching for a specific target object in natural scene images. We use the validation set and select samples where a visual search is needed to successfully locate the target. We convert the ground-truth human fixation sequence on each sample to a 2D heatmap and use it as guidance during the search. Specifically, the fixation sequence is an ordered sequence of points on the image, and we convert it to a dense 2D heatmap by adding Gaussian distributions centered at each fixation point to assign scores to each pixel. Considering the order of the points in the fixation sequence, for the $i^{th}$ fixation point, we multiply a weight $\gamma^{i}$ where $ 0 < \gamma < 1$. Then we use this heatmap generated from human fixations as guidance to guide our search process and compare it with \emph{V}$^*$ in terms of search length. Interestingly, \emph{V}$^*$ algorithm can achieve similar efficiency to the human fixations (Table~\ref{Table:cocosearch}). Examples are shown in Fig~\ref{fig:cocosearch}.

\subsection{General Multimodal Benchmarks Evaluation}
To verify that adding the visual search ability does not impede the general multimodal ability, we evaluate our model on several multimodal benchmarks including MME\cite{mme}, POPE \cite{POPE}, MMBench \cite{mmbench}, SEED-Bench \cite{seed-bench}, MM-Vet \cite{MM-Vet}, and LLaVA-Bench$^{\rm W}$ \cite{llava}. For fair comparisons, we compared with the LLaVA model trained on our VQA training data. In Table~\ref{Table:general benchmark}, we show that with the visual search mechanism, the performance on the comprehensive benchmark MME is improved and the hallucination problem is alleviated on POPE. For the larger-scale benchmarks MMBench and SEED-Bench, the performance basically remains the same. There is a slight decline in performance on MM-Vet and LLaVA-Bench$^{\rm W}$. This could be attributed to their smaller scale and the use of a GPT4-based evaluation method, which may introduce more uncertainty and potential biases. Moreover, certain questions in these benchmarks trigger a visual search for targets that are items in diagrams. This often results in the model failing to locate objects accurately because it was trained on common objects. Overall, while most common multimodal benchmarks focus on large, prominent visual elements, our model, supplemented with the visual search mechanism, still upholds its general multimodal capabilities.
\vspace{-0.5em}

\section{Conclusion}
\vspace{-0.5em}
\label{sec:conclusion}
We introduce the SEAL MLLM framework, featuring the LLM-guided visual search algorithm \emph{V}$^*$ for accurate visual grounding in high-resolution images. Our new benchmark \emph{V}$^*$Bench highlights the critical role of visual search capabilities in MLLMs. At present, our visual search model is primarily tailored to natural images and common objects. To extend its applicability to document and diagram images, long-form videos, or open-world environments, additional training and new algorithm design are necessary. Moreover, exploring architectural improvements--such as integrating convolution-based models for more efficient processing of images of any resolution, could further enhance the efficiency of the search process.

\clearpage

\small
\bibliographystyle{ieeenat_fullname}
\bibliography{main}

\clearpage

\clearpage
\appendix

\section{Implementation Details}

\subsection{Data Curation for VQA LLM}
\label{appx:data_vqa_llm}
For the GQA part of the 167k VQA data, our target is to find questions where the annotated objects mentioned in the question are critical to correctly answer the question. Therefore, we first evaluate InstructBLIP on GQA questions with annotated objects in the question and only keep questions that can be correctly answered by it. Then we use the image inpainting model LaMa \cite{lama} to erase all the mentioned objects in the corresponding images and re-evaluate the InstructBLIP model with the modified images and only keep the questions that can not be correctly answered after this modification. Through these process, we have selected a subset of GQA questions where the annotated objects are important and we use them to construct our VQA data.

For the VAW object attribution part of the 167k VQA data, we create open-ended questions and binary questions about objects' attributes. For the open-ended questions, we consider attribute types including `color', `material', `hair color', `pattern', `face expression', `pose', `activity', `opaqeness', and `texture'. For the binary questions, we additionally include attribute types `state' and `optical property'. All these attribute types follow the definition from the VAW dataset. We use fixed templates for both types of questions. For the open-ended questions, the question template is ``What is the [attribute type] of the [object name]?'' and for the binary questions, the template is ``Is the [attribute type] of the [object name] [attribute value]?'' The corresponding answer to open-ended questions is ``The [attribute type] of [object name] is [attribute value].'' and the answer to binary questions is ``Yes/No, the [attribute type] of [object name] is/is not [attribute value].'' Besides, we use the same strategy as the GQA part to filter the questions with the InstructBLIP model and the object erasing process.

For the additional 40K data created from LLaVA-80K instruction tuning data, we first extract all the noun phrases in the questions/instructions of the LLaVA-80K data. Then we choose noun phrases that are matched with the object category names defined by COCO. Note that we augment some original category names with more common synonyms (\eg add `man' and `woman' for the `person' category). Then we check whether there exist annotated instances of this category in this corresponding image. If so, we keep this sample and use the annotated instances together with their bounding box information as the targets objects for our training.

\subsection{Data Curation for Visual Search Model}
\label{appx:data_visual_search}
The training data of our visual search model includes two parts. The first part is the detection and segmentation data and the second part is the VQA data which includes possible locations QAs and LLaVA-80K instruction tuning data.

The COCO-Stuff \cite{cocostuff}, LVIS-PACO part \cite{paco}, refCOCO(+/g) \cite{refcoco,refcocog}, and refCLEF \cite{refcoco} datasets are used as both detection and segmentation data. Objects365 v2 \cite{objects365} and GoldG \cite{mdetr} datasets are only used as detection data.

The textual contextual cues of our visual search model are in the form of possible location expressions about the target objects. So we construct (image, question, answer) pairs about objects' possible locations. We randomly sample a subset of images from COCO2017. For each image, we randomly sample 2 objects that are absent in this image but appear in the five images that are most similar to it (based on CLIP embedding). The question is always asking ``What is the most likely position of [object]''. Then we provide the image information (5 captions and a list of existing objects) to GPT-3.5 and ask it to provide the possible location of the absent objects and use its response as the answer. The complete prompt is shown in Table~\ref{tab:prompt_conversation_possible_location}.

\subsection{Model Training}
\label{appx:model_training}
For the VQA LLM, we use the Vicuna-7b-1.3 \cite{vicuna} as the language model. Following the common practice of current MLLMs, the training process has two stages, a feature alignment stage and an instruction tuning stage. For the alignment stage, the linear layer projection module and the resampler projection module are separately trained along with a frozen language model and vision encoder on the subset of image-text pairs from the 558K LAION-CC-SBU subset used in LLaVA. For the instruction tuning stage, we use the constructed 387k data to train the LLM along with the projection modules, and only the vision encoder is frozen in this stage.

For the alignment stage, the linear projection module and the resampler projection module are separately trained with batch size 256. The linear projection module is trained for 1 epoch with learning rate $10^{-3}$ and the resampler module is trained for 5 epochs with learning rate $2\times10^{-4}$.  For the instruction tuning stage, the model is trained for 3 epochs with learning rate $2\times10^{-5}$ and batch size 128. To reduce the computational cost, during the training, we use the linear projection module to project the search target's feature only when there is just one object, otherwise, we use the linear projection module to project the global image feature and use the resampler for the searched objects. The exact input sequence for the LLM constructed from the VWM is:

\begin{tcolorbox}[enhanced]
\small
\texttt{<Image> \\
Additional visual information to focus on: \\ 
\{Target Object 1's Name\} <Object> at location [x1, y1, x2, y2]; \\
\{Target Object 2's Name\} <Object> at location [x1, y1, x2, y2]; \\
$\cdots$ \\
Question}
\end{tcolorbox}
Here \texttt{<Image>} is the feature tokens of the image and \texttt{<Object>} is the feature tokens of the target object stored in the VWM.

For the visual search model, we adopt the LLaVA-7B-v1.1 as the MLLM. We use the OWL-ViT-B-16's vision encoder as the image backbone. The $D_{cl}$ is trained with segmentation loss which consists of binary cross-entropy loss and DICE loss. During inference, the logits output from the $D_{cl}$ is used as the search cue heatmap. The $D_{tl}$ is trained with set prediction loss similar to DETR \cite{DETR} with focal loss \cite{focal} for coordinates regression. The whole model is trained for 100K steps with batch size 64 and learning rate $10^{-4}$. The sampling ratio of general detection/segmentation datasets (Objects365 v2, COCO-Stuff, LVIS-PACO), referring detection/segmentation datasets (refCOCO, refCOCO+, refCOCOg, refCLEF, GoldG), and VQA data is 15:8:15.

During training, the pre-trained MLLM is trained with LoRA \cite{lora} with the word embeddings layer being trainable. The image encoder of the localization module and the coordinates MLP in $D_{tl}$ are frozen. The confidence score MLP and the $D_{cl}$ are trainable.

\subsection{Visual Search Process}
\label{appx:search_process}
During the search process, we first set a relatively higher threshold and terminate when a target is located with a confidence score higher than this threshold. If the whole search process is completed without any target being found, we adjust the threshold to a lower value and find the target with the highest confidence score during the whole search process and accept it if its confidence score passes the adjusted threshold. In the scenario where the visual search is needed to find a certain target, finding all instances in a high-resolution image requires scanning the whole image exhaustively, making the search strategy less meaningful. Therefore, our current visual search process focuses on finding a single target object instead of locating all targets exhaustively. However, if we successfully locate multiple targets directly on the global image without the need for further search, we add all of them to the VWM. When evaluated on \emph{V}$^*$Bench, the target with the highest confidence score is accepted as the searched target if no target passes the threshold during the search process. In our implementation, the higher threshold is set to 0.5 and the lower threshold is set to 0.3. And the threshold $\delta$ is set to $max(3.0, 6.0\times0.7^{l})$, where $l$ is the image sub-dividing level.

To let the MLLM in the visual search model generate the search cue heatmap corresponding to the contextual cue, we extract the noun phrases in the textual contextual cue which is a possible location expression, and prompt the MLLM to locate the phrase and output the heatmap corresponding to this contextual cue.

\section{\emph{V}$^*$Bench Examples and Two Special Subsets}
\label{appx:benchmark}
\begin{figure}
\centering
\includegraphics[width=1.0\linewidth]{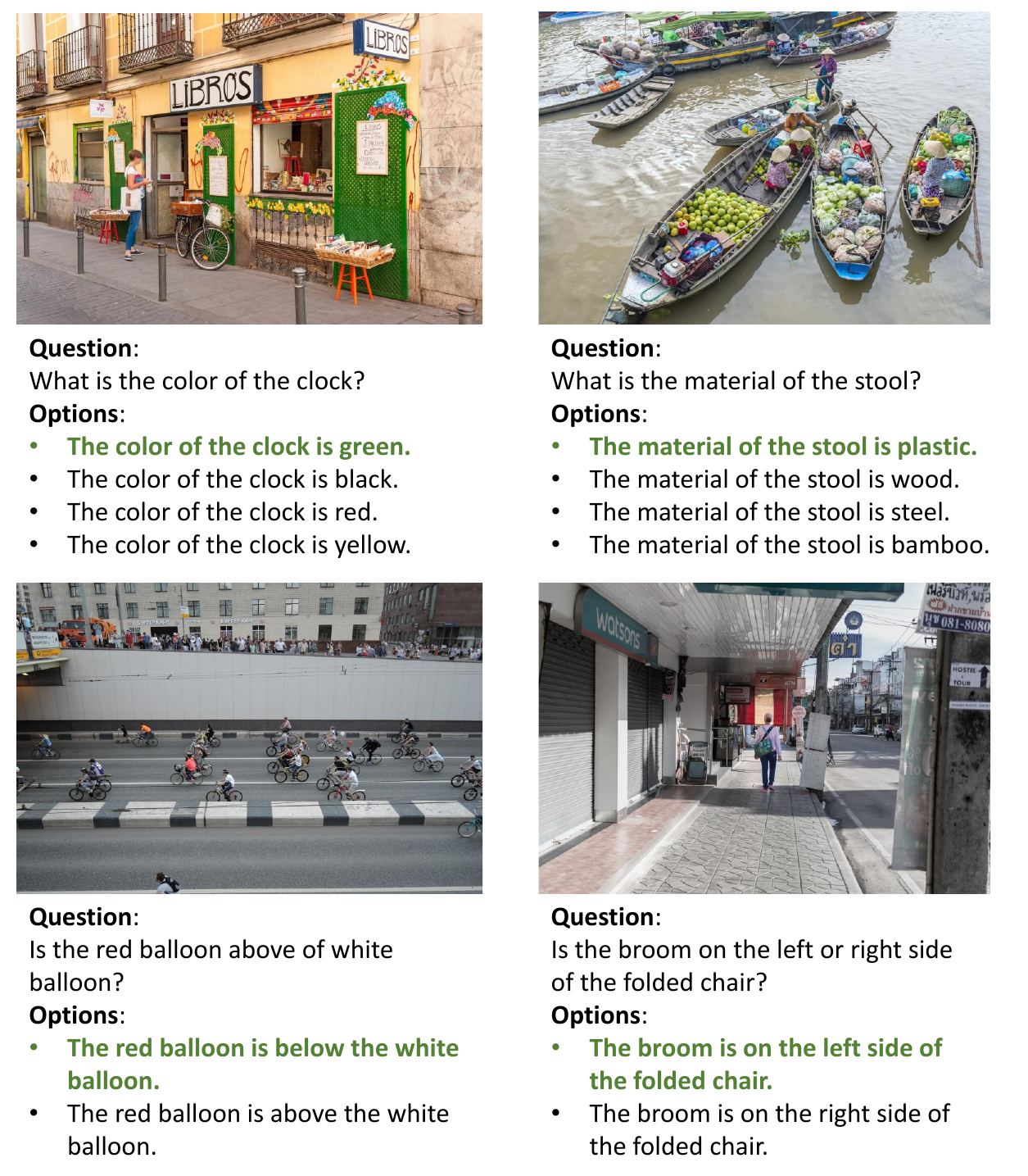}
\caption{Examples of the \emph{V}$^*$Benchmark. The top row belongs to the attribute recognition task while the bottom row belongs to the spatial relationship reasoning task. The correct option is in green.}
\label{fig:benchmark_regular}
\end{figure}

\begin{figure}
\centering
\includegraphics[width=0.8\linewidth]{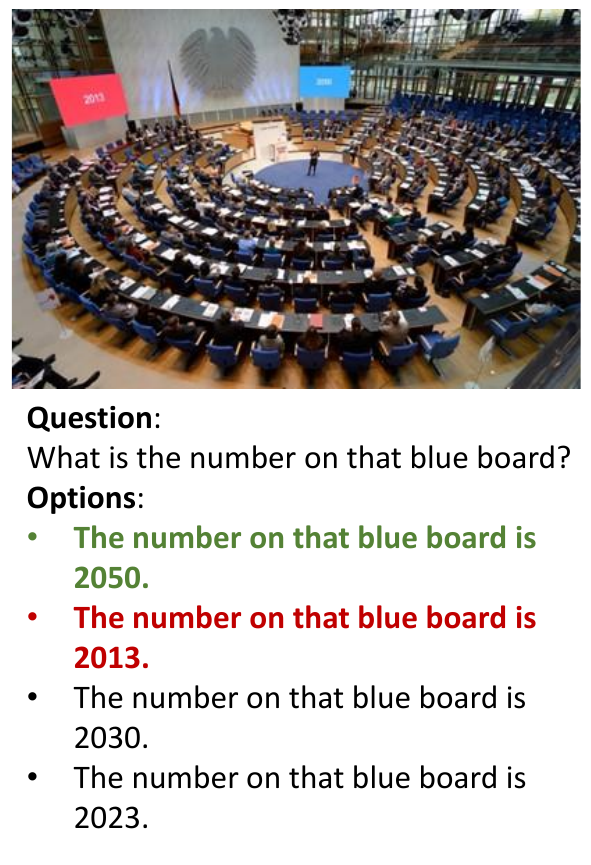}
\caption{An example on which the MM-React system with an external OCR detection model fails. The correct option is in green and the option chosen by MM-React is in red.}
\label{fig:benchmark_special}
\end{figure}

\begin{figure}
    \centering
    \includegraphics[width=0.9\linewidth]{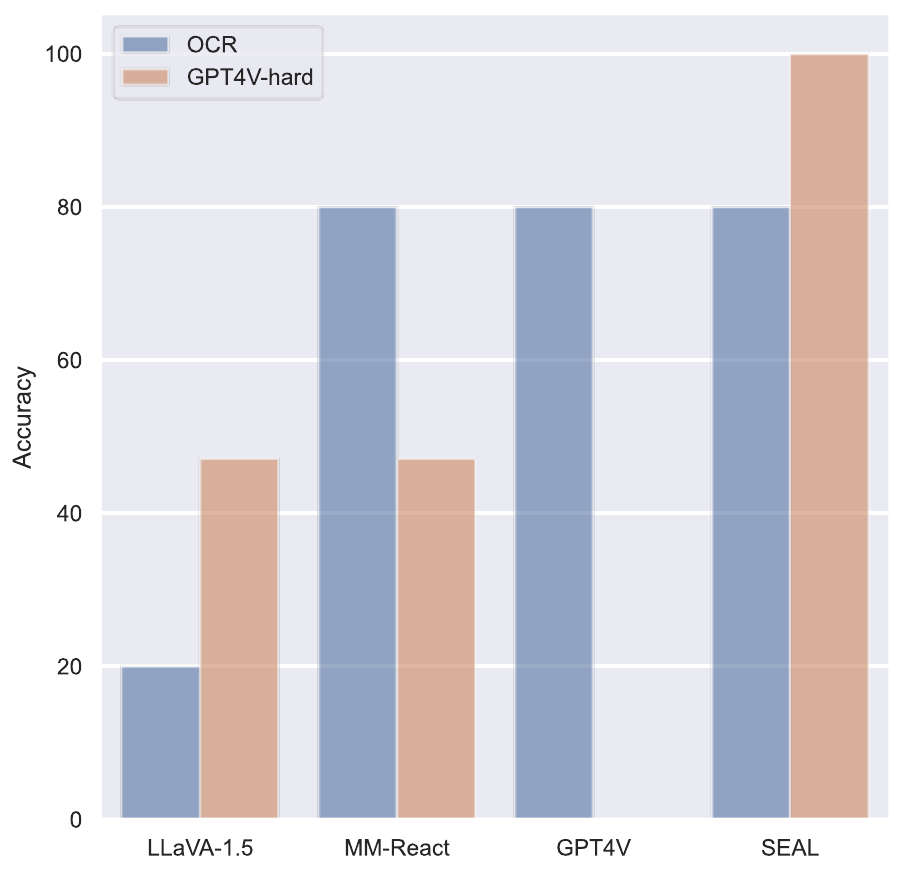}
    \caption{Comparison between SEAL and top MLLM systems on the OCR and GPT4V-hard sub-tasks.}
    \label{fig:barplot}
\end{figure}

Some examples of our proposed \emph{V}$^*$Bench are shown in Fig~\ref{fig:benchmark_regular}. Besides the regular attribute recognition and spatial relationship reasoning sub-tasks, we additionally collect two special subsets and add them to our \emph{V}$^*$Bench for exploratory study. The first subset contains 30 VQA samples which require the model to recognize and understand textual characters or digits on certain objects in the image and we denote this subset as OCR. To better expose the problem of current MLLMs and even the leading multimodal system GPT-4V, we collect 17 samples on which GPT-4V would fail but our simple model with visual search mechanism success and denote them as GPT-4V-hard. We also evaluate LLaVA-1.5, MM-REACT, GPT-4V, and SEAL on these two subsets, and the results are shown in Fig~\ref{fig:barplot}. As the MM-React system relies on the external OCR detection model and GPT-4V also likely has an OCR module, their performance on the OCR task is decent. However, for MM-React, the external OCR model detects all the texts in the image and provides them to the LLM in a bottom-up manner. Therefore, it is easy to choose the correct option merely based on the detected texts in the scene when the image only contains a few texts. When there are multiple text contents in the image or the question needs the model to fully understand the context of the text or its location, the system would fail. An example is shown in Fig~\ref{fig:benchmark_special}.

We also provide the image sources of examples in Figure 2 of the main text below:

Row-1, Column-1: \href{https://robbreport.com/travel/hotels/fao-schwarz-hotel-suite-opens-in-new-york-citys-conrad-hotel-midtown-2878934/}{Web Source}

Row-1, Column-2: Japanese animated film \emph{Children Who Chase Lost Voices}~\cite{childrenwhochaselostvoices2011}

Row-1, Column-3 \href{https://www.firstresponsecleaning.ca/commons-signs-of-hoarding/}{Web Source}

Row-1, Column-4:
\href{https://www.pexels.com/photo/new-york-city-at-dusk-18529821/}{Web Source}

Row-2, Column-1:
SR-RAW dataset~\cite{SR-RAW}

Row-2, Column-2:
Personal Photo in Orlando Disney

Row-2, Column-3:
\href{https://depositphotos.com/cn/editorial/warriors-stephen-curry-takes-free-throw-shoot-26124903.html}{Web Source}

Row-2, Column-4:
SR-RAW dataset~\cite{SR-RAW}

Row-3, Column-1:
SR-RAW dataset~\cite{SR-RAW}

Row-3, Column-2:
\href{https://www.reddit.com/r/battlestations/comments/9708ni/my_cluttered_battlestation_how_do_you_guys_keep/}{Web Source}

Row-3, Column-3:
SAM dataset~\cite{SAM}

Row-3, Column-4:
SR-RAW dataset~\cite{SR-RAW}

And the image source of the example in Figure 1 of the main text is the Japanese animated film \emph{Weathering with You}~ \cite{weatheringwithyou2019}

\section{Learning Spatial Relationship from Coordinates}
\begin{figure}
    \centering
    \includegraphics[width=\linewidth]{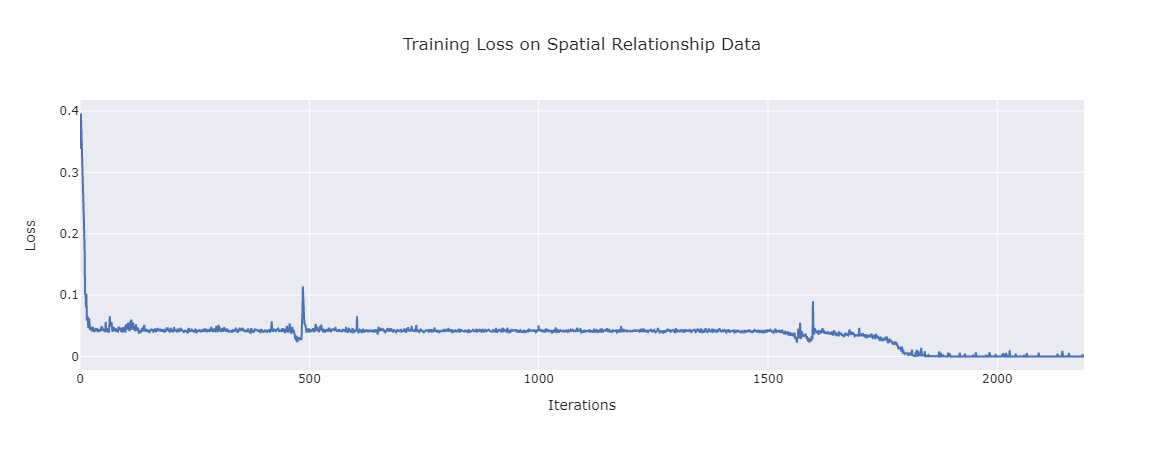}
    \caption{The loss curve of training on spatial relationship VQA data. The ``grokking'' happens after a certain number of optimization steps.}
    \label{fig:loss}
\end{figure}

We provide the numerical coordinates of the search targets to the VQA LLM as the spatial information about the searched targets. We find that though it seems intuitive and simple to recognize the relative spatial relationship with the coordinates, it is not trivial for the VQA LLM to understand the numerical coordinates and understand the spatial relationship between search targets by comparing their coordinates. We conduct additional experiments to train the VQA LLM only on the constructed 46K spatial relationship related VQA data and the loss curve is shown in Fig\ref{fig:loss}. We can see that instead of gradually decreasing, the loss suddenly drops to 0 after a certain number of optimization steps, suggesting that the model has learned to correctly compare the numerical coordinates for determining spatial relationships. As this kind of grokking needs a certain amount of optimization steps, one might need to improve the ratio of the spatial relationship related VQA data when mixing it with a larger amount of general multimodal instruction tuning data (\eg training data for LLaVA-1.5 \cite{llava1.5}) to ensure the model can correctly understand the numerical coordinates.

\begin{table*}[h!]\centering

\begin{minipage}{1.9\columnwidth}\vspace{0mm}    \centering
\begin{tcolorbox} 
    \centering
    \small
     \hspace{-6mm}
    \begin{tabular}{p{0.99\columnwidth}}
\begin{minipage}{0.99\columnwidth}\vspace{0mm}
""" \\
You are an AI visual assistant that can analyze a single image. You receive five captions, each describing the same image you are observing. And you will also be given a list of objects which are in the image. \\
Captions: \\
\{\} \\
Objects: \{\} \\
You will be given two target objects, and your task is to use your common sense knowledge and the information about the image to describe the most possible location of the given target objects in the image. Your answer must be a short expression describing the possible location referring to some other existing entities in the image. Answer concisely in less than 10 words. \\
Examples: \\
Target Object: bird \\
Most Possible Location: in the sky \\
Target Object: flag \\
Most Possible Location: on the roof of the building \\
\#\#\# \\
Target Object 1: \{\} \\
Target Object 2: \{\} \\
Output Format \\
Most Possible Location of Target Object 1: \\
Most Possible Location of Target Object 2: \\
"""
\end{minipage}
    \end{tabular}
\end{tcolorbox}
\caption{Prompt for GPT-3.5 to generate possible location expression of an object which is absent in the image.}
\label{tab:prompt_conversation_possible_location}
\end{minipage}
\end{table*}

\begin{table*}[h!]\centering

\begin{minipage}{1.9\columnwidth}\vspace{0mm}    \centering
\begin{tcolorbox} 
    \centering
    \small
     \hspace{-6mm}
    \begin{tabular}{p{0.99\columnwidth}}
\begin{minipage}{0.99\columnwidth}\vspace{0mm}
""" \\
You are an AI visual assistant that can analyze a single image. You receive five captions, each describing the same image you are observing. \\
Captions:\\
\{\}\\
You will be given 1 target object and you need to imagine certain attributes of it using your imagination and assume it is in the image.\\
You need to ask 2 questions about the provided target
object for each of the following question types: \\
Type 1: certain object attributes \\
Type 2: relative positions between objects \\
Type 3: interactions between objects \\
Your question could also involve other objects or provided information about the image. Do not ask questions about the existence of the objects. Your questions should not contain any uncertainty about the presence of the target object. Make sure each question involves the target object provided below. Ask short and natural questions of less than 20 words. \\
Target Object: \\
\{\}\\
\#\#\# \\
Output format: \\
Type 1 \\
Question 1: \\
Question 2: \\
Type 2 \\
Question 1: \\
Question 2: \\
Type 3 \\
Question 1: \\
Question 2: \\
"""
\end{minipage}
    \end{tabular}
\end{tcolorbox}
\caption{Prompt for GPT-3.5 to generate question-answer pairs about one target object which is absent in the image.}
\label{tab:prompt_conversation_negative_single}
\end{minipage}
\end{table*}

\begin{table*}[h!]\centering

\begin{minipage}{1.9\columnwidth}\vspace{0mm}    \centering
\begin{tcolorbox} 
    \centering
    \small
     \hspace{-6mm}
    \begin{tabular}{p{0.99\columnwidth}}
\begin{minipage}{0.99\columnwidth}\vspace{0mm}
""" \\
You are an AI visual assistant that can analyze a single image. You receive five captions, each describing the same image you are observing. \\
Captions: \\
\{\} \\
You will be given 2 target objects and you need to imagine certain attributes of them using your imagination and assume they are in the image. \\
You need to ask 2 questions around the provided target objects for each of the following question types: \\
Type 1: direct attribute questions about the two target objects or simple logical comparison of the attributes \\
Type 2: positions of the 2 target objects or relative positions between them or other objects \\
Type 3: interactions between the two target objects while other objects might be used as reference \\
Target Objects: \\
\{\}  \\
Your question could also involve other objects or provided information about the image. Do not ask questions about the existence of the objects. Your questions should not contain any uncertainty about the presence of the target objects. Ask short and natural questions of less than 20 words. Each question must contain both the \{\} and the \{\}. \\
\#\#\# \\
Output format: \\
Type 1 \\
Question 1: \\
Question 2: \\
Type 2 \\
Question 1: \\
Question 2: \\
Type 3 \\
Question 1: \\
Question 2: \\
"""
\end{minipage}
    \end{tabular}
\end{tcolorbox}
\caption{Prompt for GPT-3.5 to generate question-answer pairs about two target objects which are absent in the image.}
\label{tab:prompt_conversation_negative_two}
\end{minipage}
\end{table*}

\begin{table*}[h!]\centering

\begin{minipage}{1.9\columnwidth}\vspace{0mm}    \centering
\begin{tcolorbox} 
    \centering
    \small
     \hspace{-6mm}
    \begin{tabular}{p{0.99\columnwidth}}
\begin{minipage}{0.99\columnwidth}\vspace{0mm}
""" \\
Assume there is an object of type \{\} in an image, you need to come up with two visual questions asking about the visual details of the \{\}. Make sure the questions are so detailed that it is very hard to answer if the \{\} is very small in the image. Do not ask about the existence of the object. \\
Examples: \\
Object: shirt; Question: what is the text printed on the shirt? \\
Object: cup; Question: Does the cup have a dotted pattern? \\
Ask 2 reasonable questions about the \{\} and each question should be less than 20 words. \\
Object: \{\}  \\
Question 1: \\
Question 2: \\
"""
\end{minipage}
    \end{tabular}
\end{tcolorbox}
\caption{Prompt for GPT-3.5 to generate question-answer pairs about the details of a small object in the image.}
\label{tab:prompt_conversation_detail}
\end{minipage}
\end{table*}

\end{document}